# Automatic Diagnosis of Schizophrenia and Attention Deficit Hyperactivity Disorder in rs-fMRI Modality using Convolutional Autoencoder Model and Interval Type-2 Fuzzy Regression

Afshin Shoeibi[1,*], Navid Ghassemi[2], Marjane Khodatars[3], Parisa Moridian[4], Abbas Khosravi[5], Assef Zare[6], Juan M. Gorriz[7], Amir Hossein Chale-Chale[8], Ali Khadem[8], U. Rajendra Acharya[9,10,11]

1. Faculty of Electrical Engineering, FPGA Lab, K. N. Toosi University of Technology, Tehran, Iran.
2. Computer Engineering Department, Ferdowsi University of Mashhad, Mashhad, Iran.
3. Dept. of Medical Engineering, Mashhad Branch, Islamic Azad University, Mashhad, Iran.
4. Faculty of Engineering, Science and Research Branch, Islamic Azad University, Tehran, Iran.
5. Institute for Intelligent Systems Research and Innovations (IISRI), Deakin University, Geelong, Australia.
6. Faculty of Electrical Engineering, Gonabad Branch, Islamic Azad University, Gonabad, Iran;
7. Department of Signal Theory, Networking and Communications, Universidad de Granada, Spain.
8. Faculty of Electrical Engineering, K. N. Toosi University of Technology, Tehran, Iran.
9. Ngee Ann Polytechnic, Singapore 599489, Singapore.
10. Dept. of Biomedical Informatics and Medical Engineering, Asia University, Taichung, Taiwan.
11. Dept. of Biomedical Engineering, School of Science and Technology, Singapore University of Social Sciences, Singapore.

* Correspondence: Afshin Shoeibi (Afshin.shoeibi@gmail.com)

**Abstract**
Nowadays, many people worldwide suffer from brain disorders, and their health is in danger. So far, numerous methods have been proposed for the diagnosis of Schizophrenia (SZ) and attention deficit hyperactivity disorder (ADHD), among which functional magnetic resonance imaging (fMRI) modalities are known as a popular method among physicians. This paper presents an SZ and ADHD intelligent detection method of resting-state fMRI (rs-fMRI) modality using a new deep learning method. The University of California Los Angeles dataset, which contains the rs-fMRI modalities of SZ and ADHD patients, has been used for experiments. The FMRIB software library toolbox first performed preprocessing on rs-fMRI data. Then, a convolutional Autoencoder model with the proposed number of layers is used to extract features from rs-fMRI data. In the classification step, a new fuzzy method called interval type-2 fuzzy regression (IT2FR) is introduced and then optimized by genetic algorithm, particle swarm optimization, and gray wolf optimization (GWO) techniques. Also, the results of IT2FR methods are compared with multilayer perceptron, k-nearest neighbors, support vector machine, random forest, and decision tree, and adaptive neuro-fuzzy inference system methods. The experiment results show that the IT2FR method with the GWO optimization algorithm has achieved satisfactory results compared to other classifier methods. Finally, the proposed classification technique was able to provide 72.71% accuracy.

## 1. Introduction

The human brain consists of a complex and large network of neurons responsible for controlling and monitoring all parts of the body [1-2]. The brain network has different areas that are constantly connected. Many neurons in different parts of the brain are connected and coordinated to perform any function in the body, creating complex brain patterns [3-4]. Whenever connectivity is not well-connected in different areas of the brain, it can cause changes in the function of the brain and disorders such as attention deficit hyperactivity (ADHD) [5], Schizophrenia (SZ) [6-7], epilepsy [8-10], and Parkinson's disease (PD) [11]. Initially, the diagnosis of various brain disorders was based on the DSM. The experience of a specialist physician plays a critical role in diagnosing the type of brain disorder [12-13]. Therefore, examining and diagnosing the type of brain disorder is not possible for all physicians

[14]. For example, ADHD has some of the same symptoms as other brain disorders, such as PD and SZ, and is difficult for some specialists to diagnose [15-16].

To overcome these challenges, functional neuroimaging modalities have been developed to diagnose a variety of brain disorders that are highly popular with physicians [17-18]. Functional neuroimaging modalities include electroencephalography (EEG) [19], functional magnetic resonance imaging (fMRI) [20], positron emission tomography (PET) [21], single-photon emission computed tomography (SPECT) [22], magnetoencephalography (MEG) [23] and functional near-infrared spectroscopy (fNIRS) [24]. FMRI modalities are the most important non-invasive techniques for assessing brain function during brain disorders [20]. fMRI includes task fMRI (T-fMRI) and resting-state fMRI (rs-fMRI) modalities and shows spatial resolution in the brain, making fMRI modalities popular for examining functional connectivity in different parts of the brain [25]. The fMRI data contains a 4-dimensional tensor so that the 3D volume of the brain is segmented into small areas, and the activity of each area is recorded for a certain period of time [25]. In this modality, the two brain regions are functionally related if they have simultaneous functional activity [25]. FMRI modalities have been used to analyze brain connectivity patterns in diagnosing various brain disorders by physicians and have yielded promising results [26-28].

In recent years, studies have shown that functional connectivity analysis based on fMRI modalities plays a significant role in the diagnosis of brain disorders such as SZ [6], Alzheimer's disease (AD) [29], epilepsy [30-31], ADHD [5], and BD [32]. Brain disorders alter functional connectivity in the brain, which fMRI modalities can see. However, diagnosis of brain disorders with fMRI modalities is a difficult and time-consuming task for physicians [25]. For example, ADHD is a developmental disorder with some of the same symptoms as other brain disorders, such as SZ [7], making it difficult for physicians to diagnose the type of brain disorder. The main issue with only relying on physicians for the diagnosis of mental disorders such as ADHD and Schizophrenia is that it is time-consuming and subjective measures may influence the final assessments of subjects. Hence, the possibility of misdiagnosis may increase. Application of machine learning approaches for the diagnosis, can overcome these limitations and provide fast, and accurate classification.

In some cases, both ADHD and SZ disorders have been observed simultaneously in patients, but it does not mean that one disease necessarily causes the other [106-107]. Other research has shown that people with close relatives with ADHD are more likely to develop SZ [108-109]. Dopamine is known as one of the causes of ADHD and SZ. Dopamine, as a neurotransmitter, may affect attention, concentration, pleasure, happiness, and motivation [110]. In addition, other researchers have shown that perinatal risk factors may be involved in both diseases [111]. According to references [112-113], there may be an overlap between ADHD and SZ. ADHD and SZ disorders usually have a detrimental effect on memory and attention [114]. Similar to ADHD, SZ sufferers may experience challenges such as clarity of thinking. On the other hand, there are distinct differences between ADHD and SZ disorders regarding symptoms, diagnosis, and treatment [115-116]. To address these challenges, in recent years, artificial intelligence (AI) methods have been used to diagnose various brain disorders from fMRI data, and researchers are trying to create working tools to diagnose these diseases [33-34]. In this paper, the authors aim to present new deep learning (DL)-based method for the simultaneous diagnosis of SZ and ADHD, so a brief description of SZ and ADHD is provided below.

SZ is a severe and chronic mental disorder characterized by impaired thinking, perception, and behavior. In addition, SZ causes psychosis, which is associated with significant disability [7]. DSM-5 is a guide for assessing and diagnosing mental disorders [12-13]. According to the DSM-5, two or more of the symptoms of delusions, hallucinations, disorganized speech, grossly disorganized or catatonic behavior, and negative symptoms must frequently occur over one month [12-13]. Numerous factors in SZ include genetic factors, environmental factors, and brain structure and function. While there is still no cure for SZ, research has led to more innovative and safer treatments. Researchers are also trying to diagnose the disease and identify its causes by studying and conducting research and using fMRI neuroimaging techniques to study the function of the brain.

ADHD is a neurodevelopmental disorder that is very common in children ages 3 and 6 [35-36]. ADHD includes three types of inattention, hyperactivity-impulsivity, and combination [35-36]. ADHD always causes many challenges for children, including forgetting simple things, talking non-stop, being distracted, etc. [35]. ADHD has different symptoms for each person. Therefore, its accurate diagnosis is challenging for specialist doctors [36].

Nowadays, a lot of research is being done to diagnose brain disorders using DL techniques [37-39]. The DL models are inspired by the human nervous system, consisting of several interconnected layers of processing units with nonlinear activation functions called artificial neurons [40-42]. One of the advantages of DL networks is their ability to directly process large raw data or data with little preprocessing on them to automatically extract the best features for the task [40-42].

So far, various researches have been done on diagnosing SZ [7] and ADHD [43] using fMRI modalities and DL techniques. The main purpose of these pieces of research is to increase the performance for SZ or ADHD detection using MRI modalities. In addition, the researchers hope to be able to provide a working DL-based tool for the diagnosis of SZ and ADHD in the future. Further details of these studies are summarized in Table (6). However, among the presented studies, simultaneous diagnosis of SZ and ADHD from fMRI data using DL techniques has not been performed.

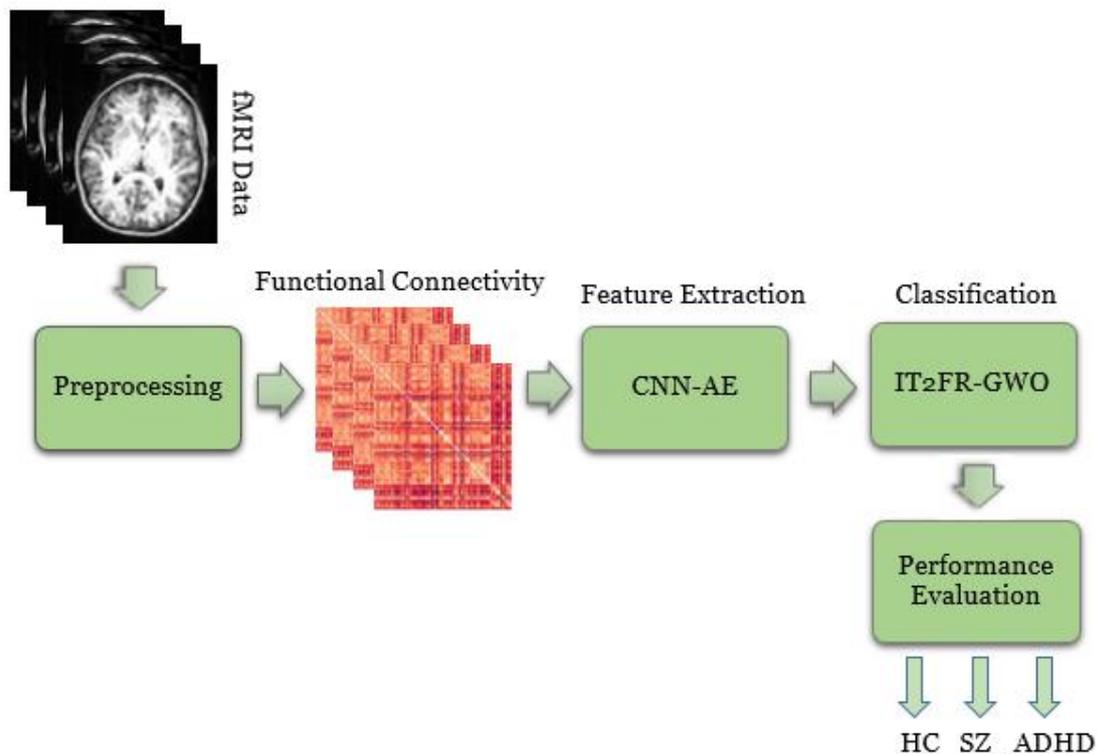

Fig. 1. Block diagram of the proposed method.

This paper presents a new SZ and ADHD detection method in rs-fMRI modality using 2D convolutional Autoencoder (CNN-AE) and interval type-2 fuzzy regression (IT2FR)- gray wolf optimization (GWO) methods. First, rs-fMRI data from the University of California Los Angeles (UCLA) dataset were used for SZ and ADHD detection [44]. The FMRIB software library (FSL) [45] toolbox is used to preprocess rs-fMRI data. The preprocessing performed on rs-fMRI data includes various steps that are discussed in detail in the next section of this manuscript. Afterward, a CNN-AE model with the proposed layers is used to extract features from preprocessed fMRI data. Finally, the classification of features is done by different classifier methods, namely, decision tree (DT) [57], multilayer perceptron (MLP) [46], k-nearest neighbors (KNN) [47], support vector machine (SVM) [48], random forest (RF) [49], adaptive neuro-fuzzy inference system (ANFIS) [50], ANFIS-GA (genetic algorithm), ANFIS-PSO (particle swarm optimization), ANFIS-GWO, IT2FR-GA, IT2FR-PSO, and IT2FR-GWO. In this step, the IT2FR technique is introduced as a pioneering work and optimized using GA [51], PSO [52], and WGO [53] optimization algorithms. Also, combining the IT2FR with WGO is another novelty of this paper. In the following, other sections of the paper are described. The second section introduces the proposed method of SZ detection from rs-fMRI data and the CNN-AE model. The third section is dedicated to the introduction of evaluation parameters. In the fourth section, the test results of the proposed method are presented. Finally, the discussion, conclusions, and future work are presented in the fifth section.

## 2. Proposed Method

The proposed methods for SZ and ADHD detection from rs-fMRI modalities are introduced in this section. In Figure (1), the steps of the proposed method are shown in the form of a diagram. First, the UCLA dataset containing rs-fMRI modalities was used. In the second step, preprocessing is performed on rs-fMRI data using FSL software. This step is based on standard preprocessing methods for rs-fMRI modalities. The following feature extraction method is based on the CNN-AE architecture with the suggested number of layers. Finally, various classification algorithms including MLP, KNN, SVM, RF, ANFIS, ANFIS-GA, ANFIS-PSO, ANFIS-GWO, IT2FR-GA, IT2FR-PSO, and IT2FR-GWO are used. In the following, each of these sections is presented.

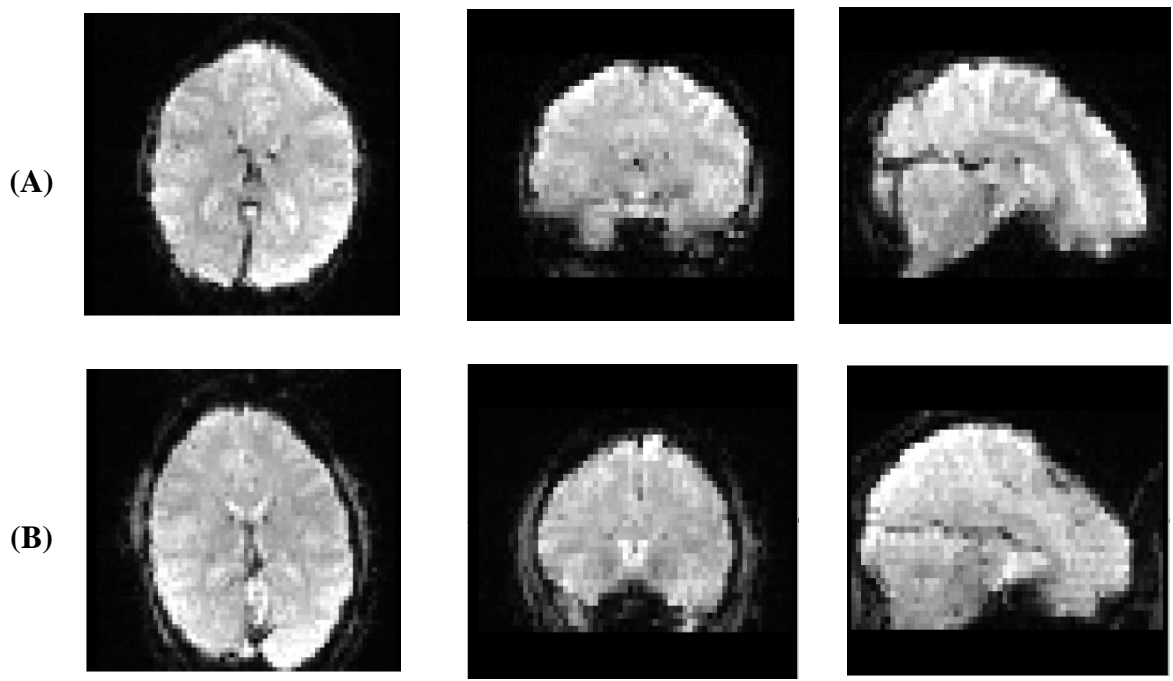

Fig. 2. Sample rs-fMRI data from HC subjects (row A) and patients (row B).

### 2.1. Dataset

This paper used the UCLA dataset for our experiments [44]. This dataset has different versions; the first version is presented in reference [134]. The dataset includes fMRI data from 138 healthy subjects, patients with Schizophrenia (58), Bipolar disorder (49), and ADHD (45). Additional information about the dataset (sex, age and etc.) can be found in [134] [44]. In this paper, we used the healthy control (60 subjects), SZ (58 subjects), and ADHD (45 subjects) classes. We selected 60 HC subjects randomly from 138 subjects to make the dataset balanced between different classes. Figure (2) depicts some samples of rs-fMRI data from the UCLA dataset.

Images were acquired using a 3T Siemens Trio scanner. The rs-fMRI data with 304 seconds (eyes-open) was collected from each subject using T2*-weighted echo-planar imaging (EPI) pulse sequence with parameters: axial slices, number of slices=34, slice thickness=4mm, TR=2s, TE=30ms, flip angle=90°, matrix=64×64, FOV=192×192 mm$^2$. Also, a T1-weighted high-resolution anatomical scan (MPRAGE) was acquired from each subject with the following parameters: sagittal slices, number of slices=176, slice thickness=1mm, TR=1.9s, TE=2.26ms, matrix=256×256, FOV=250×250 mm$^2$. The traditional statistical analyses (i.e., ANOVA and t-test) are conducted on demographic information and functional connectivity. Table (1) provides the details about the data used in this work with statistical analysis.

Table 1. Details about the data used in this work with statistical analysis.

| **Variables** | HC | SZ | ADHD | P Value |
|---|---|---|---|---|
| No. | 60 | 58 | 45 | - |
| Age (mean ± s.t.d) | 31.59 ± 8.78 | 36.46 ± 8.78 | 32.05 ± 10.27 | 0.019 |
| Hand Score (mean ± s.t.d)** | 0.91 ± 0.15 | 0.95 ± 0.09 | 0.91 ± 0.10 | 0.23 |
| Sex (Male : Female)*** | 29 : 23 | 38 : 12 | 21 : 19 | 0.03 |

*The P value was computed using a one-way Anova analysis
**A measure between 0 and 1. 1: Completely right-handed, 0: completely left-handed
***P Value computed using chi Square test

A one-way ANOVA (p-value <0.0005, Uncorrected) has been conducted on the functional connectivities. Table (2) shows the summary of one-way ANOVA results obtained for the functional connectivity.

Table 2. Summary of one-way ANOVA results obtained for the functional connectivity.

| Region Pairs | P value |
|---|---|
| Right Temporal Pole--Right Frontal Medial Cortex | 0.0004 |
| Right Frontal Medial Cortex--Right Frontal Orbital Cortex | 0.0001 |
| Right Subcallosal Cortex--Right Frontal Orbital Cortex | 0.0002 |
| Left Central Opercular Cortex--Left Thalamus | 0.0002 |
| Left Central Opercular Cortex--Left Brain-Stem | 0.0002 |
| Left Central Opercular Cortex--Right Brain-Stem | 0.0001 |

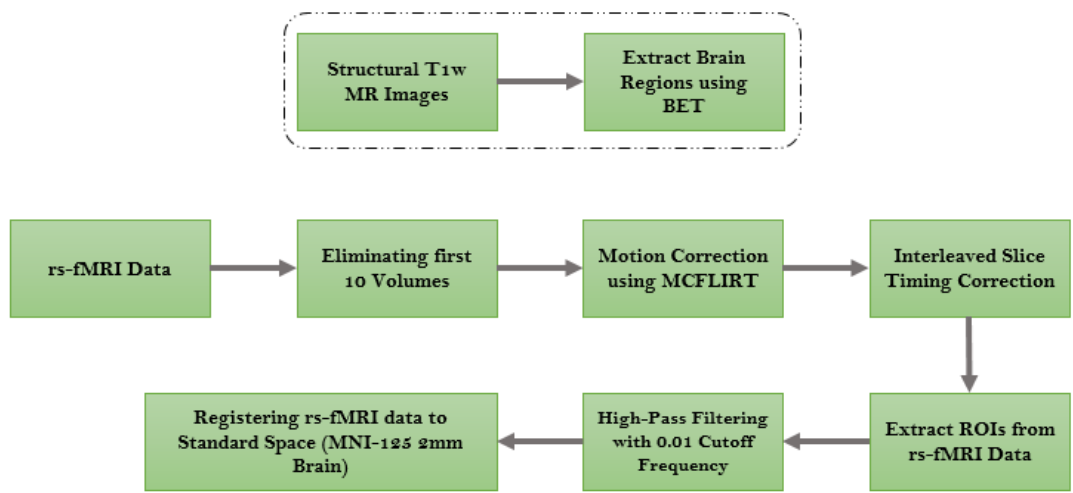

Fig. 3. Overview of the preprocessing steps.

## 2.2. Preprocessing

All preprocessing steps were conducted using the FSL software [45]. A general overview of the preprocessing steps can be seen in figure (3) and are as follow:
- Eliminating non-brain regions from the T1 high-resolution images using BET
- Eliminating the first 10 volume images to allow for magnetization equilibrium
- Applying a temporal Gaussian high-pass filter with a cut-off frequency of 0.01 HZ

- Applying motion correction using FSLs MCFLIRT
- Applying interleaved slice timing correction
- Applying a Gaussian Smoothing with FWHM = 5mm for noise reduction
- Registering all images to standard space (MNI152_T1_2mm)

The outcome of the main preprocessing steps (filtering and registration) on a sample image can be seen in figures (4) and (5).

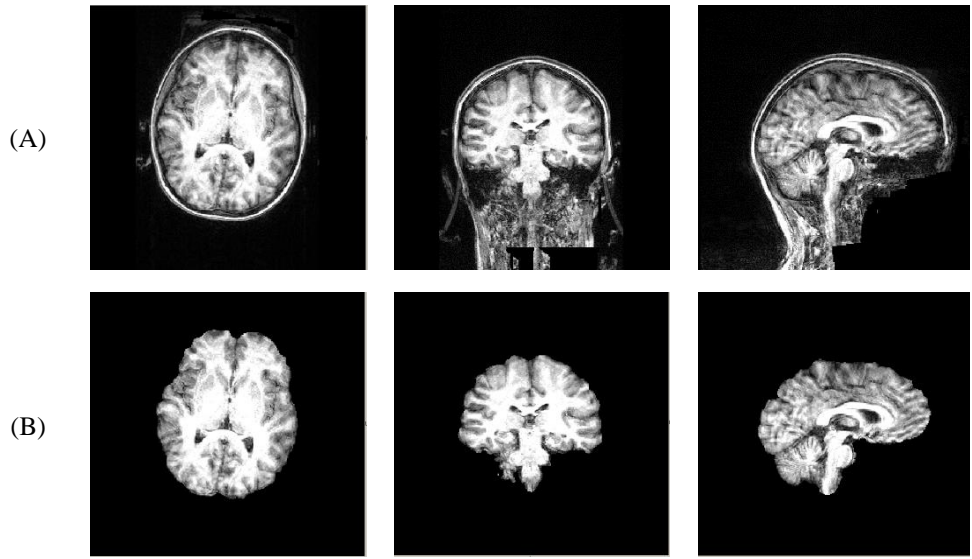

(A)

(B)

Fig. 4. Sample images before and after applying BET on raw T1-images.

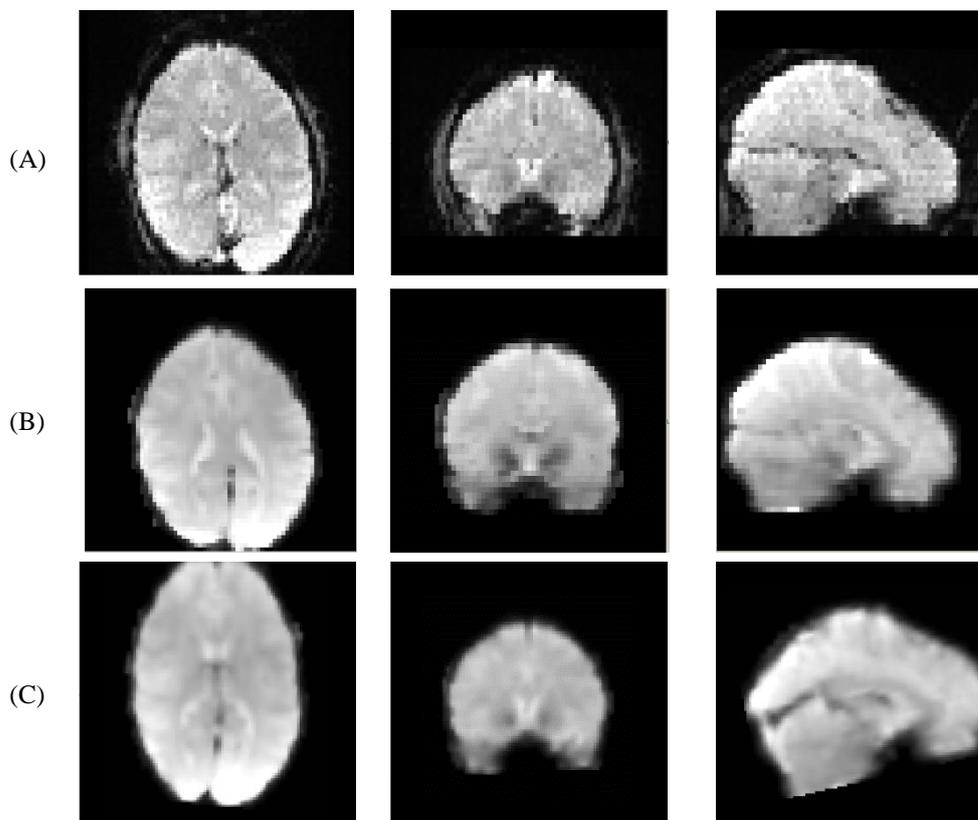

(A)

(B)

(C)

Fig. 5. Sample images after three main preprocessing steps. Row (A): raw rs-fMRI data. Row (B): Raw rs-fMRI data after brain extraction, slice timing correction, and filtering. Row (C): Registered Image to standard space.

### 2.2.1. Connectivity

In fMRI modalities, functional connectivity expresses the temporal correlation between the time series of different brain regions [117-118]. In other words, the presence of statistical dependencies between two neurophysiological data is investigated by functional connectivity [117-118]. Assuming that the data fit Gaussian assumptions, then second-order dependencies are discussed and include covariance's or correlations [117-118]. For non-Gaussian processes, higher-order dependencies are also investigated by some methods such as independent component analysis (ICA) [119]. The most important methods of functional connectivity are correlation [120], coherence [121], mutual information [122], transfer entropy [123], directed coherence [124], Granger causality [125], generalized synchronization [126], and Bayes net [127].

In this paper. the Pearson correlation coefficient was used to estimate the functional connectivity between regions. Regions of interest (ROI) were defined according to the Harvard-Oxford cortical and sub-cortical (lateralized) atlas. Based on this atlas, a total of 118 ROIs were considered. Next, regional time series were obtained by averaging the time series of all voxels inside each of the 118 ROIs. Finally, the Pearson correlation coefficient between the time series was used to extract the connectivity matrices for each subject. In conclusion, a connectivity matrix with a shape of 118×118 is obtained. Figures (6) to (8) demonstrate connectivity heat maps from healthy control (HC) and SZ subjects.

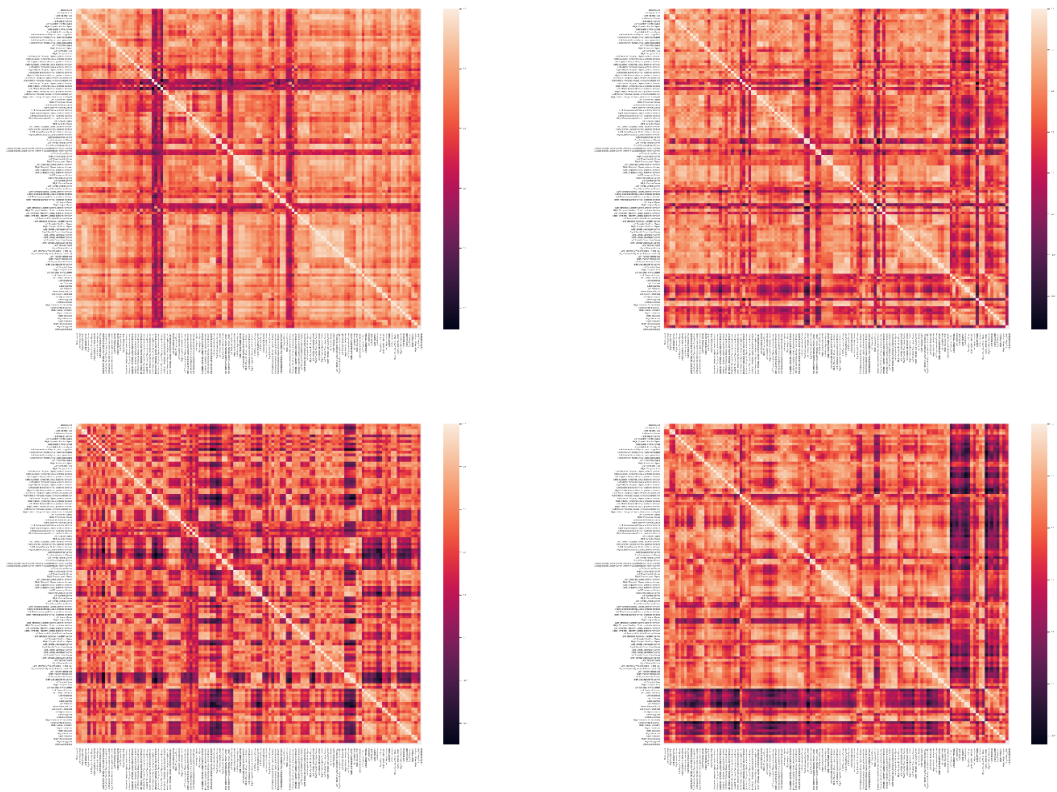

Fig. 6: Sample correlation matrices obtained for HC subjects.

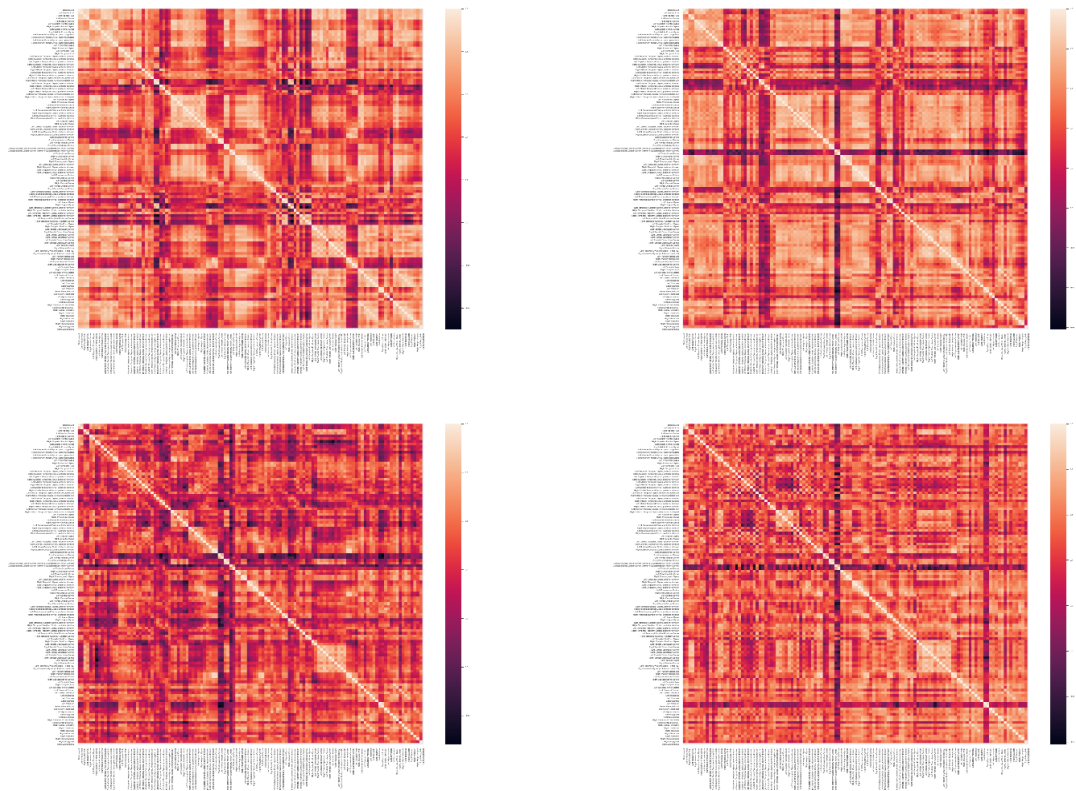

Fig. 7. Sample correlation matrices obtained for SZ subjects.

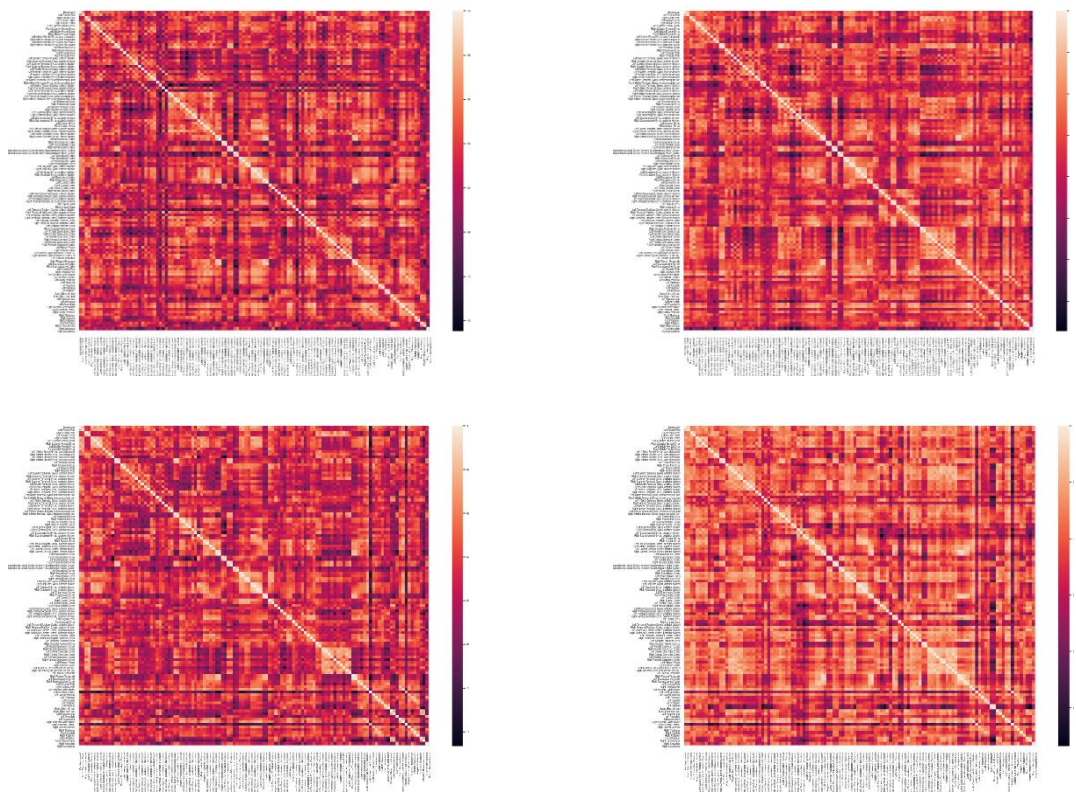

Fig. 8. Sample correlation matrices obtained for ADHD subjects.

## 2.3. Proposed CNN-AE Model

The introduction of DL techniques in various fields, including medicine [25-30], has made significant progress, and valuable results have been achieved in detecting different diseases [33-34]. So far, several models of DL techniques have been proposed, with different training schemes, including supervised, semi-supervised, and unsupervised methods [54-55]. Autoencoders (AEs) are an important class of DLs trained unsupervised and are widely used for feature extraction [55-56]. AE architectures consist of two important parts: decoder and encoder [55].

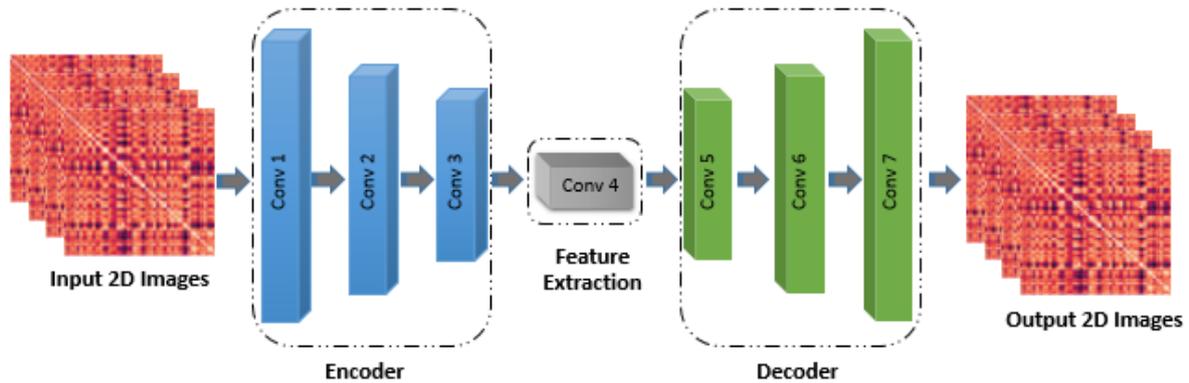

Fig. 9. Proposed CNN-AE model for diagnosis of SZ from rs-fMRI modality.

AEs have different architectures [55], and the idea of this paper is based on the CNN-AE technique with the proposed layers. In the proposed model, the connectivity matrices are fed to the input of the encoder section. Then, it passes through the convolutional layers to reach the last layer of the encoder. Afterwards, this data is entered into the decoder, which aims to perform data reconstruction. This paper uses seven convolutional layers in the encoder and decoder; three of these are in the encoder (along with activation function and followed by a max pooling), and four are in the decoder (the first three are followed by an upsampling layer, while the last one is used to generate images with the same shape as input). The output of the forth convolution layer (before upsampling) is used for feature extraction. This structure was picked by empirically testing a few alternatives. Figure (9) and Table (3) detail the proposed CNN-AE model.

Table 3. Details for the proposed CNN-AE model.

| Architecture | Layers | Output Shape | | Kernel Size | Stride | Param | Activation |
|---|---|---|---|---|---|---|---|
| | | Width | Depth | | | | |
| | Input Layer | (118, 118) | 1 | -- | -- | 0 | -- |
| | Conv2D | (118, 118) | 32 | 32 | 1 | 320 | ReLU |
| | Max Pooling | (59, 59) | 32 | -- | 1 | 0 | -- |
| Encoder | Conv2D | (59, 59) | 32 | 32 | 1 | 9248 | ReLU |
| | Max Pooling | (30, 30) | 32 | -- | 1 | 0 | -- |
| | Conv2D | (30, 30) | 1 | 32 | 1 | 289 | ReLU |
| | Max Pooling | (15, 15) | 1 | -- | 1 | 0 | -- |
| | Conv2D | (15, 15) | 1 | 32 | 1 | 10 | ReLU |
| | Upsampling | (30, 30) | 1 | -- | 1 | 0 | -- |
| | Conv2D | (30, 30) | 32 | 32 | 1 | 320 | ReLU |
| Decoder | Upsampling | (60, 60) | 32 | -- | 1 | 0 | -- |
| | Conv2D | (60, 60) | 32 | 32 | 1 | 9248 | ReLU |
| | Upsampling | (120, 120) | 32 | -- | 1 | 0 | -- |
| | Zero Pad | (118, 118) | 1 | -- | -- | 0 | -- |
| | Conv2D | (118, 118) | 1 | 1 | 1 | 289 | Tanh |

## 2.4. Classifiers

First, the decoder part is deleted to use an autoencoder for feature extraction. Next, a dense layer followed by a SoftMax classifier is inserted after the encoder, and the encoder is fine-tuned. Afterward, the representations from the last convolutional layer (last layer of the encoder) are extracted as features for each image. Figure (10) shows the overall structure of the fine-tuned network. After performing these steps, a representation for each image is generated which can be fed to different classifier methods. In this paper, the used classification algorithms consist of MLP, KNN, SVM, RF, ANFIS, ANFIS-GA, ANFIS-PSO, ANFIS-GWO, IT2FR-GA, IT2FR-PSO, and IT2FR-GWO. In the following, the details of classifier algorithms are presented.

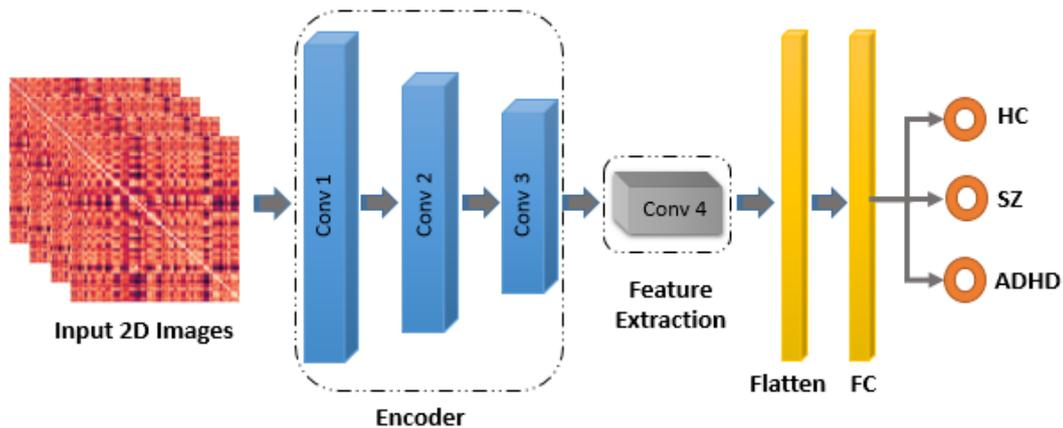

Fig. 10. AE-based Classifier block diagram.

### 2.4.1. Machine learning classifiers

To compare the results of the proposed method and show its superiority, several classification algorithms are picked and evaluated on the dataset. Decision trees (DT) [57], KNN [47], MLP [46], SVM [48] and Random Forest [49] are used. These methods have different complexities, and each has its strength and weaknesses; for example, random forest is fast and robust against outliers but tends to fail where features are highly entangled, whereas SVM with the proper kernel can solve those problems, but its performance decreases for massive datasets. Moreover, some of these methods have hyper-parameters, k in KNN and kernel in SVM; these values are picked as three and RBF, respectively.

### 2.4.2. ANFIS Models

Nowadays, various applications use fuzzy systems, including medical data classification [50]. The ANFIS model is one of the most popular classification methods, which combines fuzzy logic and adaptive network. More information about the ANFIS method is given in [50]. In this work, the Genfis-3 function is used to implement ANFIS based on the FCM clustering algorithm. Furthermore, this function uses the Gaussian membership function for input data. Finally, the training step is based on the hybrid method. In [50], ANFIS model training with intelligent optimization methods is presented, and we have done the same in this research.

### 2.4.3. Improving Type-2 Fuzzy Regression Based on TSK

Linear regression (LR) has been extensively utilized in many applications, such as the medical one [58-59]. According to the definition of LR, there will be a linear relationship between independent variables and the dependent variable. LR-based approaches are among well-known and widely used machine learning tools (ML) [60].

Due to uncertainty in medical data and the capability of Fuzzy Logic Systems (FLSs) in modelling uncertainties, Type-1 and Type-2 Fuzzy Regression methods have been proposed, which have more potential and efficiency to outperform LR techniques.

Moreover, taking advantage of additional degrees of freedom provided by the footprint of uncertainty (FOU), Interval Type-2 Fuzzy Logic Systems (IT2 FLSs) are able to effectively handle the high levels of uncertainty in comparison with T1 FLS [61].

The following definitions will be presented before analyzing the Interval Type-2 Fuzzy Regression (IT2 FR).

***Definition 1[62]:*** The T2FS is represented by a type-2 membership function $\mu_{\tilde{A}}(x,u)$ as follows:

$$\tilde{A} = \{((x,u), \mu_{\tilde{A}}(x,u)) | \forall x \in X, \forall u \in J_x \subseteq [0,1]\} \quad (1)$$

where $x$ is the primary variable in $X$, $u$ is the second variable which has the domain $J_x \subseteq [0,1]$, and the amplitude of $\mu_{\tilde{A}}(x,u)$ is known as the second grade. The T2FS can also denoted as:

$$\tilde{A} = \int_{x \in X} \int_{\mu \in J_x} \mu_{\tilde{A}}(x,u)/(x,u) \quad (2)$$

***Definition 2:*** The footprint of uncertainty (FOU) of a T2FS $\tilde{A}$ is defined as

$$\tilde{A} = \{(x,u) | x \in X, u \in [\underline{u}_{\tilde{A}}, \overline{u}_{\tilde{A}}]\} \quad (3)$$

***Definition 3[62]:*** Due to the complexity and abstraction of T2FS, IT2FS is defined by setting all $\mu_{\tilde{A}}(x,u) = 1$, which can be shown as

$$\tilde{A} = \int_{x \in X} \int_{\mu \in J_x} 1/(x,u) \quad (4)$$

The $i$th IF-THEN rule of T2 FLSs is expressed by

$$\text{if } x_1 \text{ is } \tilde{A}_1^i \text{ and } x_2 \text{ is } \tilde{A}_2^i \text{ and}, \dots, x_n \text{ is } \tilde{A}_n^i, \text{then } y^i = \tilde{G}^i, \quad i = 1, \dots, M \quad (5)$$

where $\tilde{A}_j^i (j = 1, \dots n)$ are antecedent type-2 sets, $\tilde{G}^i$ are consequent type-2 sets. Further studies about the structure of T2FLSs and their applications are proposed in [63]. Figure (11) demonstrates that IT2 FS could adopt the Gaussian MFs for its premise parts.

***Definition 4 [62]:*** Gaussian IT2FS is a special case of IT2FS which has a fixed mean $(m)$, and an uncertain standard deviation that takes on values in $[\sigma_1, \sigma_2]$. The lower membership function (LMF) $\underline{\mu}_{\tilde{A}}(x)$ and Upper membership function (UMF) $\overline{\mu}_{\tilde{A}}(x)$ of the Gaussian IT2FS are expressed as follows:

$$\underline{\mu}_{\tilde{A}}(x) = e^{-\frac{1}{2}\left(\frac{x - m(\tilde{A})}{\sigma_1(\tilde{A})}\right)^2} \quad (6)$$

$$\overline{\mu}_{\tilde{A}}(x) = e^{-\frac{1}{2}\left(\frac{x - m(\tilde{A})}{\sigma_2(\tilde{A})}\right)^2} \quad (7)$$

Figure (12) shows LMFs and UMFs of an IT2FS with three Gaussian MFs.

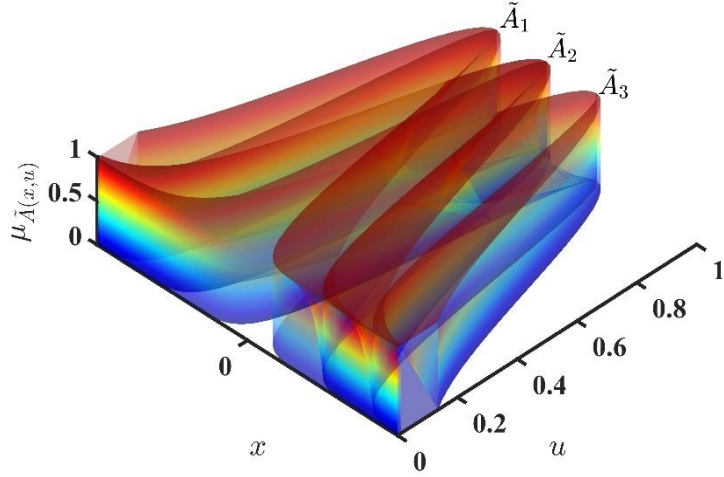

Fig. 11. Gaussian IT2F membership functions.

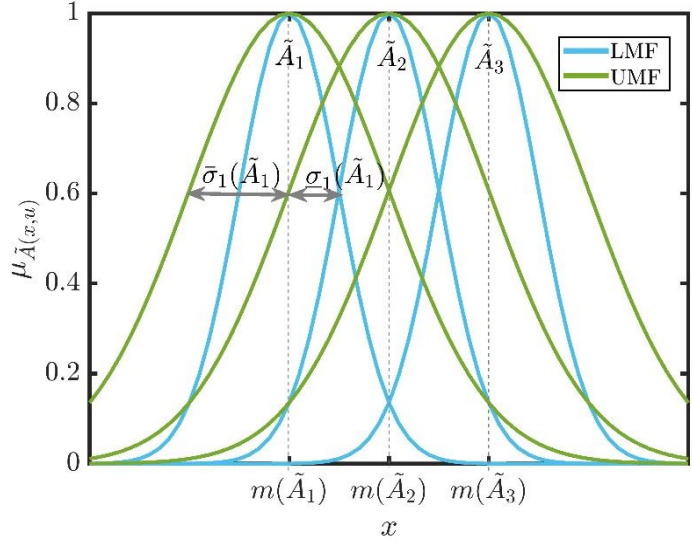

Fig. 12. Upper MFs and lower MFs of Gaussian IT2F.

When data in a regression problem follow an almost certain linear trend in a region, linear regression methods will be efficient. On the other hand, piecewise linear regression is employed if the linear trend does not exist. The number and location of breakpoints need to be detected in the estimation process. This issue raises complexity and additionally leads to an algorithm whose behavior changes due to added data. This issue could be the leading cause of repeating all calculations.

In this section, IT2 FR is developed to overcome the abovementioned difficulty. Moreover, this method will be independent of the behavior of a trend in a data set. For IT2FR, output $Y_i$ is defined by an interval type-2 fuzzy linear model in which $\tilde{A}_j$ and $X_j^{(i)}$ are fuzzy coefficients and crisp inputs:

$$\tilde{Y}_i = \tilde{A}_1 X_1^{(i)} + \cdots + \tilde{A}_n X_n^{(i)} = \sum_{j=1}^{n} \tilde{A}_j X_j^{(i)} \tag{8}$$

The above equation does not apply to data with various behavior patterns. To solve this problem, data with similar behavior will be allocated to a cluster, and then linear IT2FR will be used for each cluster. Only that particular cluster will be trained in analyzing new data if new data is placed in the existing clusters. Otherwise, a new cluster will be generated, and IT2FR associated with this cluster will be studied. As a result, the previous calculations are valid, while new behavior patterns are also investigated with minimum computational cost.

Assume that $N$ and $M$ are the number of data and clusters, respectively, for $r$th cluster with $M_r$ members, the following rule could be expressed:

$$if\ x_1^{(i)} is\ \tilde{F}_1^r\ and\ x_2^{(i)} is\ \tilde{F}_2^r\ and\ \ldots\ x_n^{(i)} is\ \tilde{F}_n^r\ ,then\ \tilde{Y}_i^r = \sum_{j=1}^n \tilde{A}_j^{r} x_j^{(i)} \tag{9}$$

$$i = 1, \ldots, M_r, r = 1,2,\ldots,M, \sum_{r=1}^M M_r = N$$

In fact, one rule is defined for each cluster in which $\tilde{F}_j^r$ are interval type-2 fuzzy sets (IT2FSs) for $r$th rule and will be determined as follows:

First, with the help of Fuzzy C-Mean clustering (FCM) method, the optimal number of clusters, the cluster members and T2Fs $\tilde{F}_j^r$ are determined.

The fuzzy regression function based on input and output variables is obtained for each cluster. Consequently, the fuzzy coefficients $\tilde{A}_j^{r}$ are calculated.

The final output is defined as follows:

$$\tilde{Y}_i = \sum_{r=1}^M h_r(x_1^{(i)}, \ldots, x_n^{(i)})\tilde{Y}_i^r = \sum_{r=1}^M \sum_{j=1}^n h_r(x_1^{(i)}, \ldots, x_n^{(i)})\tilde{A}_j^{r} x_j^{(i)} \tag{10}$$

$$= \sum_{j=1}^n \sum_{r=1}^M [h_r(x_1^{(i)}, \ldots, x_n^{(i)})\tilde{A}_j^{r}] x_j^{(i)} = \sum_{j=1}^n \tilde{A}_j^{'}(x_1^{(i)}, \ldots, x_n^{(i)}) x_j^{(i)}$$

Where

$$\tilde{A}_j^{'}(x_1^{(i)}, \ldots, x_n^{(i)}) = \sum_{r=1}^M h_r(x_1^{(i)}, \ldots, x_n^{(i)})\tilde{A}_j^{r} \tag{11}$$

$$h_r(x_1^{(i)}, \ldots, x_n^{(i)}) = \frac{T(\tilde{F}_j^r)}{\sum_{r=1}^M T(\tilde{F}_j^r)}, h_r \in (0,1) \tag{12}$$

The above relation demonstrates a nonlinear fuzzy regression dealing with data complexity. $h_r$ is a coefficient to activate $r$th rule and $T(\tilde{F}_j^r)$ is:

$$T(\tilde{F}_j^r) = \pi_{j=1}^n \mu(\tilde{F}_j^r(x_j)) \tag{13}$$

When data belongs only to a single cluster, it could obviously determine only one FR relation in the output part.

Data patterns that simultaneously belong to many clusters affect more than one FR relations. The degree of each data effectiveness could be specified by its membership grade.

Remark 1: If a cluster's members do not overlap another cluster's members, then $h_r \in \{0,1\}$, and we have exclusive fuzzy linear regression.

The output of FS $Y_i$ after type-reduction and defuzzification [64] stages will be

$$Y_i = \sum_{r=1}^M h_r(x_1^{(i)}, \ldots, x_n^{(i)}) a_j^r \in [Y_i^L, Y_i^R] \tag{14}$$

where $a_j^r$ is the center of T2FS $\tilde{A}_j^{'}$. The final output is

$$Y_i^* = \frac{Y_i^L + Y_i^R}{2} \tag{15}$$

### 2.4.4. Optimal IT2FR Classifiers

Fuzzy c-mean clustering (FCM) method [50] is the most useful approach to generating a primary fuzzy inference system (FIS) among the ANFIS and IT2FR classifiers. In FIS, direct learning methods influence the Gaussian membership functions (MFs), inputs, and outputs of FIS. This section employs PSO, GA, and GWO methods to train IT2FR and ANFIS models. Firstly, based on FCM, the proposed FR model obtains T1 and T2 Gaussian MFs, and then optimization algorithms are used to decrease the error. Finally, the objective function is defined by [50]:

$$\min_\theta Error = \frac{1}{N} \sum_{i=1}^{n} e_i^2 \tag{16}$$

Where

$$e_i = t_i - f(x_i|\theta) \tag{17}$$

and we have

$$RMSE = \sqrt{\frac{1}{n} \sum_{i=1}^{n} (t_i - y_i)^2} \tag{18}$$

where $N$ and $e_i$ are the number of FS inputs and error, respectively. $x_i$ are inputs (features) [50]. $\theta$, $n$, and $y_i$ denote the IT2FR and ANFIS method parameters, the number of samples, and the output of FS, respectively.

### A) Genetic Algorithm

The genetic algorithm (GA) is one of the most important optimization methods developed by John Holland [51]. This algorithm starts with a set of chromosomes called populations. Chromosomes are taken from a population and used as a new population. Then, the hypothesis begins with a unique random population and continues through generations [51]. The fitness of the whole population is assessed in each generation. Then, several individuals are randomly selected from the current generation (based on fitness) and modified to form a new generation. The next iteration converts the algorithm to the current generation [51]. You can find more information on GA in [51].

### B) Particle Swarm Optimization (PSO)

The PSO optimizes a problem by trying iteratively to enhance a candidate solution to an estimated quality in computational science [52]. It moves the particles in the search space using a mathematical formula over the particle's velocity and position. The local best-known position influences the particle's movement, but the best-known positions in the search space also guide them. Hence, the better positions found by other particles are also updated. This makes the swarm move toward the best solutions [52]. One of the most significant advantages of this algorithm is that it does not use the gradient of the problem being optimized, eliminating all the overheads of gradient calculation and derivative computation. Nevertheless, given the nature of the local search, it does not guarantee an optimal solution [52].

### C) Gray Wolf Optimization

The GWO algorithm was first proposed in 2014 by Mirjalili et al. [53]. GWO is a metaheuristic optimization method inspired by how gray wolves hunt [53]. This technique is in the category of swarm intelligence algorithms, and it is based on population. Gray wolves are predators and often prefer to live in groups [53]. Each group consists of 5 to 12 wolves on average in a hierarchical manner. This hierarchy is divided into four categories: alpha (group leaders), beta (advisors and orders to lower groups), delta (observers and hunters), and omega (protectors) [53]. The GWO algorithm is based on the mass hunting of wolves, which will be examined in detail below.

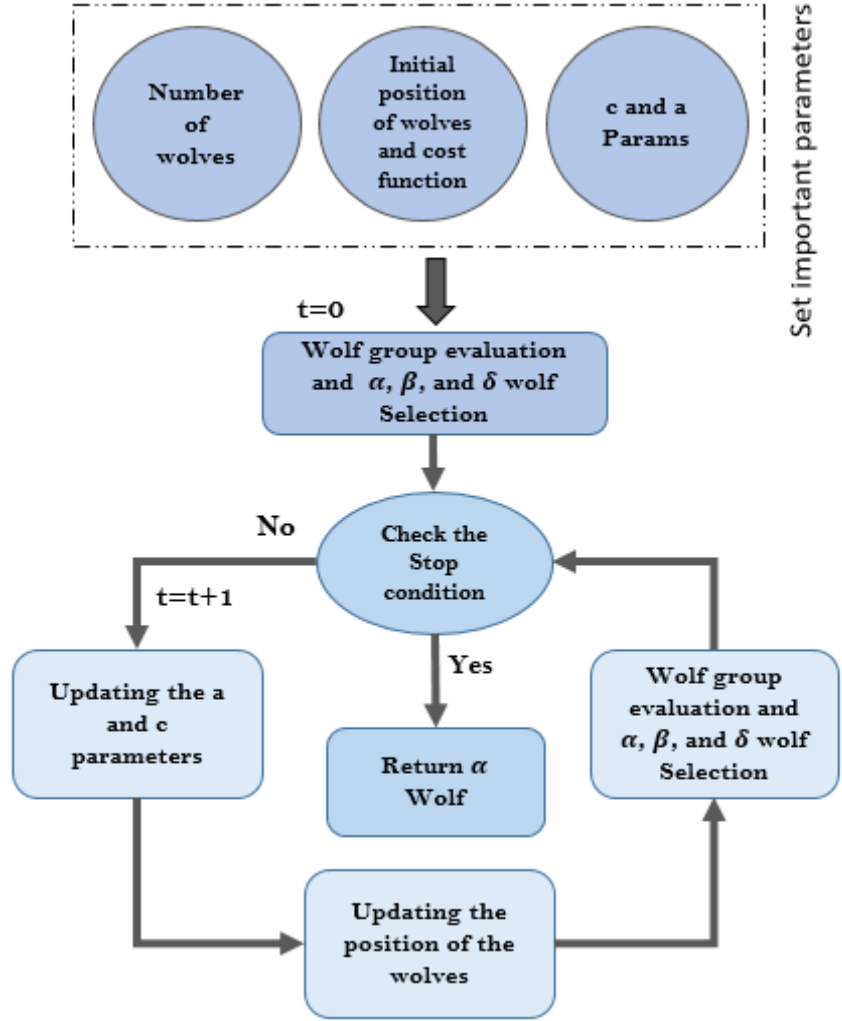

Fig. 13. Block diagram of GWO optimization method.

For the design of GWO, the most important factors include best position (α), second solution (β), third solution (δ), and other solutions (ω). For example, the mathematical equation for the siege of prey by gray wolves is defined as follows [53]:

$$X(t+1) = X_p(t) - A * D(t)$$

$$D(t) = |C * X_p(t) - X(t)|, \quad t = 1,2,\ldots, tMax$$

$$A = a(2r_1 - 1), \quad C = 2r_2, \quad r_1 \& r_2 \; are \; random \; vector \; \in [-1,1]$$

In the above relations, $X(t)$ represents the position of the gray wolf in the i-th repeat, and $X_p(t)$ represents the hunting position in the *t* repeat. Also, component α is reduced linearly from 2 to zero [53].

In the search space, we do not know $X_p(t)$, so we consider the hunting position α. Therefore, hunting equations are presented as follows [53]:

$$X_1(t) = X_\alpha(t) - A_1 * D_\alpha(t), \quad D_\alpha(t) = |C_1 * X_\alpha(t) - X(t)|$$
$$X_2(t) = X_\beta(t) - A_2 * D_\beta(t), \quad D_\beta(t) = |C_2 * X_\beta(t) - X(t)|$$
$$X_3(t) = X_\delta(t) - A_3 * D_\delta(t), \quad D_\delta(t) = |C_3 * X_\delta(t) - X(t)|$$

$$X(t+1) = \frac{(X_1(t) + X_2(t) + X_3(t))}{3}$$

Figure (13) shows the steps of the GWO algorithm. More details of the GWO method are provided in [53].

### 2.5. Statistical Metrics

This section introduces the statistical parameters for evaluating the proposed method of SZ detection from the rs-fMRI modality. In this research, the K-Fold cross-validation with k = 10 is used. Evaluation parameters include accuracy (Acc), precision (Prec), recall (Rec), and F1-Score (F1). The equations of evaluation parameters are shown below [65-66].

$$Acc = \frac{TP + TN}{FP + FN + TP + TN} \quad (19)$$

$$Prec = \frac{TP}{FP + TP} \quad (20)$$

$$Rec = \frac{TP}{TP + FN} \quad (21)$$

$$F1 = \frac{2\,TP}{2\,TP + FP + FN} \quad (22)$$

### 3. Experiment Results

This section reports the results of the proposed SZ and ADHD detection method from the rs-fMRI modality. To implement the proposed method, a PC with NVidia 1070 GPU, 16 GB RAM, and Core i7 CPU is used. The rs-fMRI preprocessing steps are performed by the FSL toolbox. Also, all the simulations were done by Scikit-learn [67] and TensorFlow 2 toolboxes [68] in Python 3.7 environment. As mentioned in the previous sections, all simulations are performed on the UCLA dataset [44], which contains rs-fMRI data from HC, BD, ADHD, and SZ subjects. Then, as explained in previous sections, preprocessing, feature extraction and classification steps are performed. For model evaluation, we used k-fold cross validation with k = 10. The authors introduced the proposed CNN-AE method, which is the paper's primary novelty. Also, the hyperparameters of the conventional ML classifiers are selected by trial and error to obtain the highest performance. In section 2-3, the details for the proposed CNN-AE model is provided. The experiment results of the proposed CNN-AE model with ML classifier techniques are shown in Table (4). It can be noted from the table that the KNN algorithm has achieved better results than other ML classifiers. The confusion matrix for the KNN model is shown in Figure (14); given that we have used k-fold cross-validation, we have averaged the confusion matrices of each fold to get a final one. Also, Figure (15) displayed the results of statistical metrics for all ML classifiers.

Table 4. Results for ML classification methods

| Methods | Acc(%) | Prec(%) | Rec(%) | F1-score (%) |
|---|---|---|---|---|
| DT | 60.90 | 64.66 | 54.83 | 64.27 |
| MLP | 57.18 | 51.678 | 52.03 | 55.63 |
| KNN | 67.72 | 65.61 | 71.94 | 68.45 |
| SVM | 66.90 | 64.57 | 64.20 | 67.70 |
| RF | 62.72 | 65.21 | 55.92 | 63.61 |

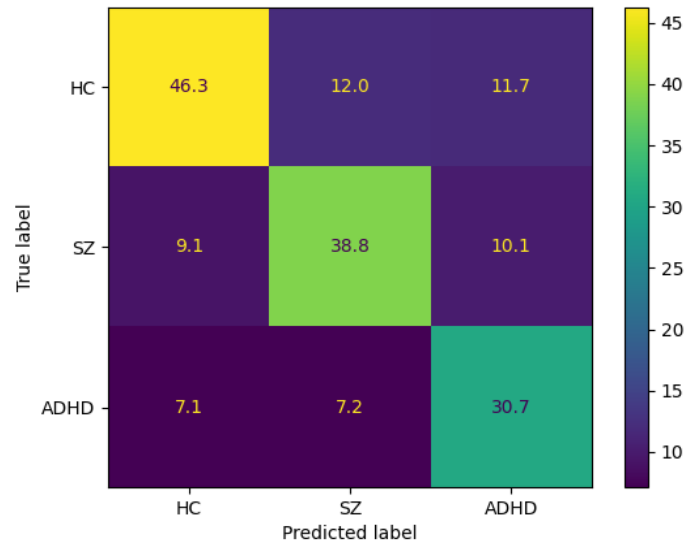

Fig. 14. Confusion matrix for KNN method.

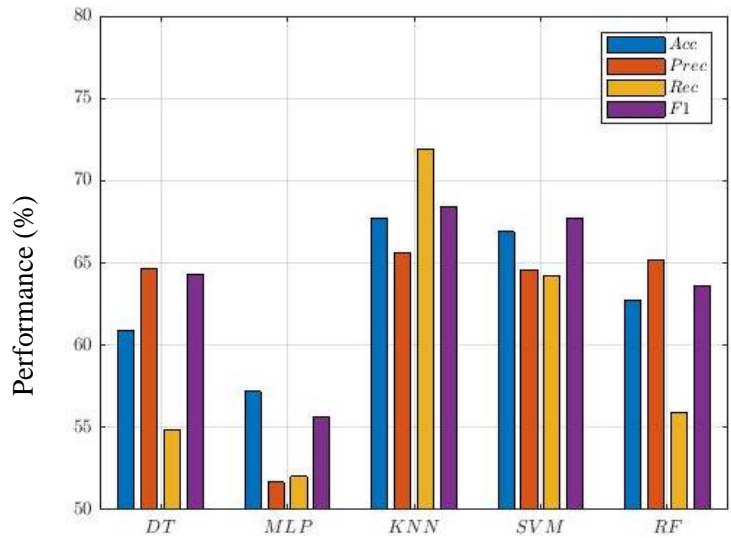

Fig. 15. Comparison of performances (%) for SZ and ADHD detection.

In the following, the experiment results of the proposed CNN-AE model with ANFIS classifier methods are presented. The ANFIS is one of the most important type-1 fuzzy systems [69]. This method is based on a Takagi-Sugeno-Kang (TSK) system [50] and has achieved successful results in various medical research [70-71]. There are several training methods for ANFIS, and the hybrid technique is used in this paper. Also, in another part of the simulations, the GA [51], PSO [52], and GWO [53] optimization algorithms were used for the training of the ANFIS model.

In different studies, the researchers have used optimization techniques in ANFIS training and achieved promising results [72-74]. To implement the ANFIS model, the FCM method is first used, which is based on Gaussian MFs [50]. Since it is a classification problem, three Gaussian MFs are provided for each input data. Figure (16) shows several Gaussian MFs for SZ and ADHD detection. Then, the hybrid method and optimization techniques expressed to train the ANFIS model were tested and compared. In Table (5), the hyper-parameters of GA [51], PSO [52], and GWO [53] algorithms are presented for ANFIS training. Finally, the results of the CNN-AE model with ANFIS, ANFIS-GA, ANFIS-PSO, and ANFIS-GWO classifier algorithms are shown in Table (6).

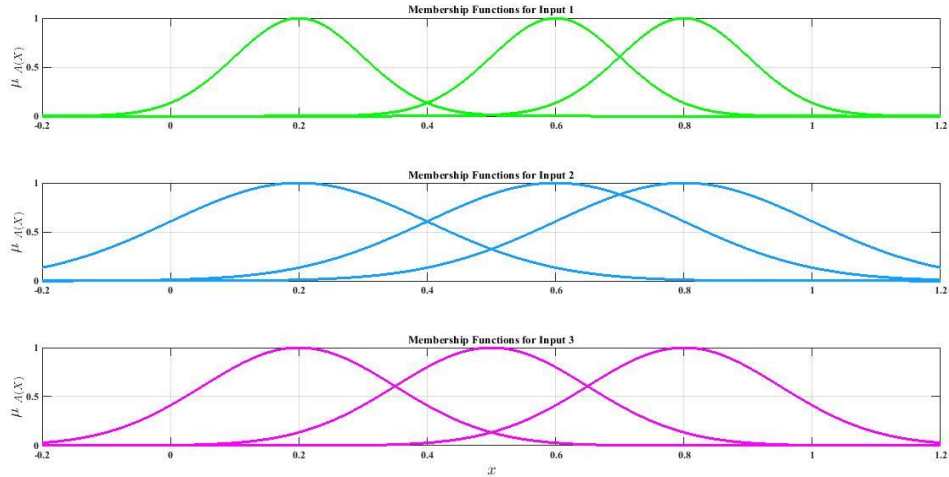

Fig. 16. Three Gaussian membership functions in ANFIS classifier.

Table 5. Hyper-parameters to optimization methods.

| PSO | GA | GWO |
|---|---|---|
| C1=2 | Selection = tournament | Search agents = 5 |
| C2=2 | Mutation rate = 0.05 | Dimension = 30 |
| W=0.2 | Crossover fraction = 0.8 | -- |
| -- | Elite Count = 5 | -- |
| N_pop=60 | | |
| Var_min = min (Feature_Matrix) | | |
| Var_max = max (Feature_Matrix) | | |
| MAX_IT=400 | | |

The experiment results in Table (6) show the superiority of the ANFIS-GWO method compared to the other ANFIS algorithms. The reason for the high accuracy of ANFIS-GWO compared to other ANFIS models is the use of the GWO method in the training step. Researchers have not used the GWO algorithm to train fuzzy systems such as ANFIS [50], which is another novelty of this paper. The confusion matrix for the ANFIS-GWO technique is displayed in Figure (17). Also, Figure (18) displays the results of statistical metrics for the ANFIS classifier.

Table 6. Results for ANFIS classifier models.

| Methods | Acc (%) | Prec (%) | Rec (%) | F1 (%) |
|---|---|---|---|---|
| ANFIS | 64.91 | 64.58 | 65.01 | 64.54 |
| ANFIS-GA | 67.05 | 66.37 | 66.31 | 66.20 |
| ANFIS-PSO | 67.22 | 67.04 | 67.48 | 66.88 |
| ANFIS-GWO | 68.15 | 67.65 | 68.17 | 67.64 |

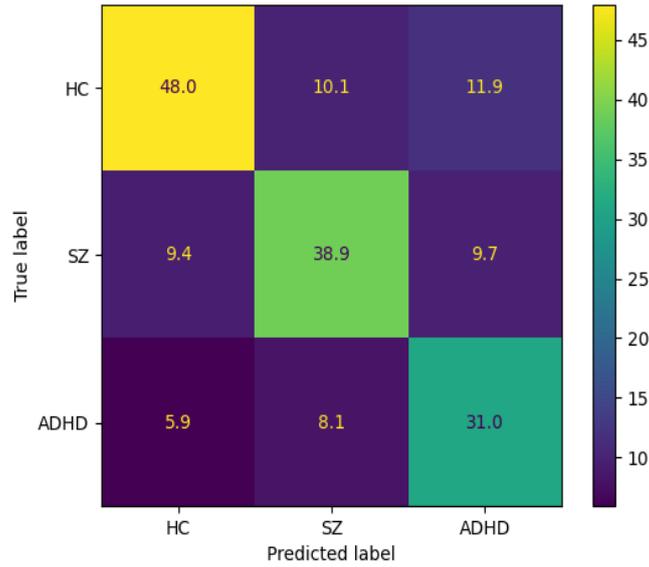

Fig. 17. Confusion matrix for ANFIS-GWO.

In the following, we present the results of the proposed CNN-AE model alongside IT2FR classifiers. The authors have proposed the IT2FR model for the first time, and it is considered the most important novelty of this paper. Similar to ANFIS, the proposed IT2FR technique is optimized by GA [51], PSO [52], and GWO [53] methods. To implement of IT2FR model, three interval type-2 Gaussian MFs are provided for each input data. Figure (19) shows several intervals of type-2 Gaussian MFs for SZ and ADHD detection. Table (7) reports the results obtained from IT2FR, IT2FR-GA, IT2FR-PSO, and IT2FR-GWO methods. According to Table (7), it is clear that the IT2FR-GWO method has achieved the most accuracy. The confusion matrix for the IT2FR-GWO method is shown in Figure (20). Also, Figure (21) displayed the results of statistical metrics for IT2RF classifiers.

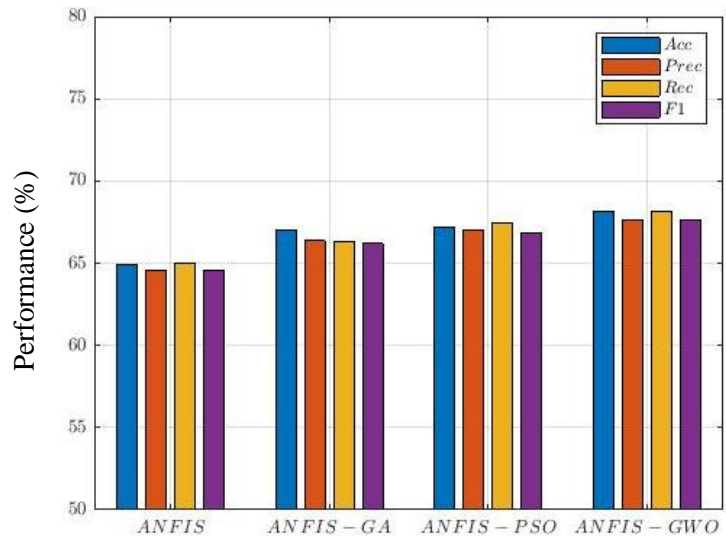

Fig. 18. Comparison of ANFIS classifiers results for SZ and ADHD detection.

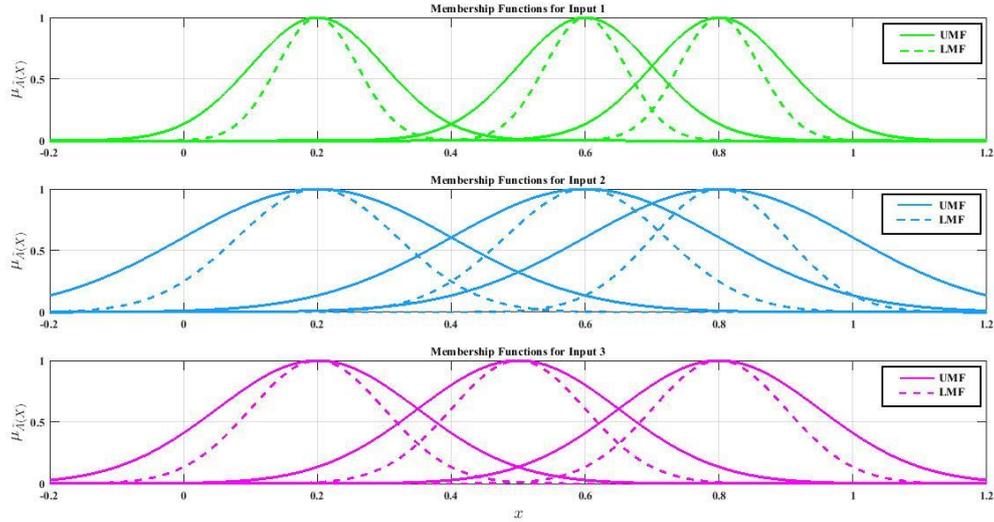

Fig. 19. Three Gaussian membership functions in IT2FR classifier.

Table 7. Results for IT2FR classifier models.

| Methods | Acc (%) | Prec (%) | Rec (%) | F1 (%) |
| --- | --- | --- | --- | --- |
| IT2FR | 68.61 | 68.29 | 68.84 | 68.27 |
| IT2FR-GA | 71.27 | 70.94 | 71.43 | 70.87 |
| IT2FR-PSO | 71.90 | 71.56 | 71.64 | 71.34 |
| IT2FR-GWO | 72.71 | 72.52 | 72.81 | 72.41 |

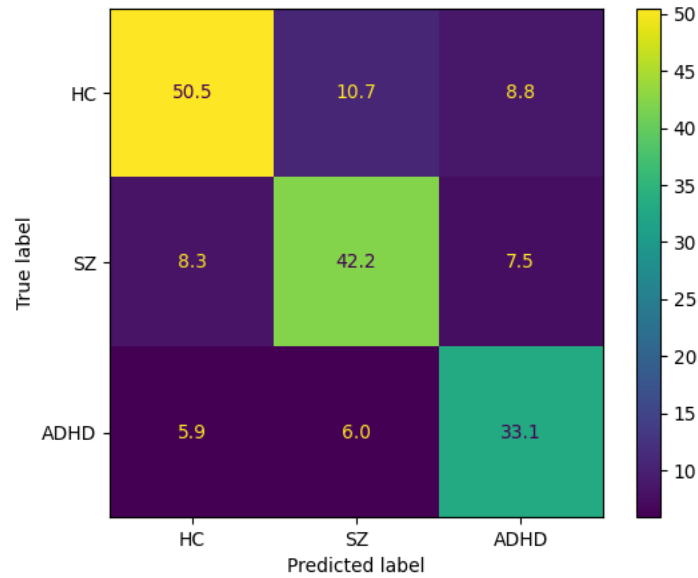

Fig. 20. Confusion matrix for IT2FR-GWO.

In this section, the results of the proposed method for the diagnosis of SZ and ADHD disorders were discussed. The proposed method of this research has three important novelties, including introducing the CNN-AE model with proposed layers, the IT2FR method, and the GWO optimization method to train the proposed IT2FR technique. In Tables (4), (6), and (7), the accuracy of different classification algorithms are shown and compared. It can be seen that the IT2FR-GWO method has achieved state-of-the-art results in comparison with other classification methods.

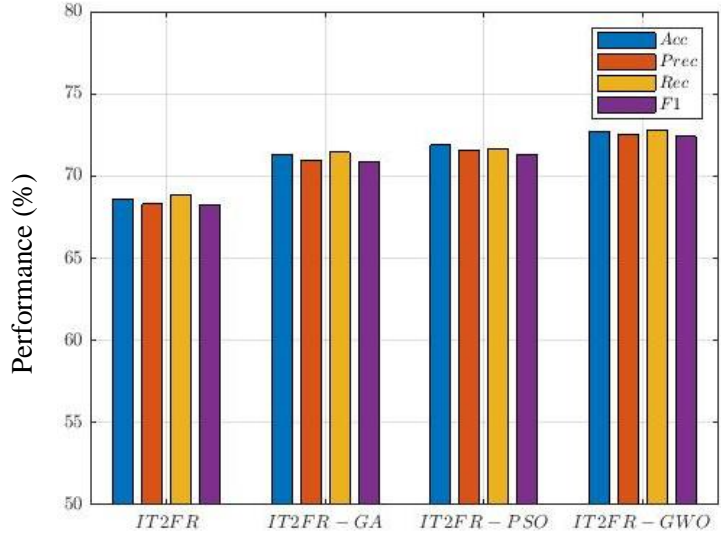
Fig. 21. Comparison of IT2RF classifiers results for SZ and ADHD detection.

## 4. Limitations of Study

This section is dedicated to the limitations of this study. The UCLA dataset has limited rs-fMRI data to diagnose various brain disorders, including SZ and ADHD. This challenge has made it still impossible to provide advanced DL models for the diagnosis of SZ and ADHD. Also, this dataset contains rs-fMRI data and does not have T-fMRI modalities from subjects with mental disorders. The UCLA dataset is used to diagnose SZ, BD, and ADHD disorders, while diagnosing brain disorders of varying severity is very important. The proposed method of this paper consists of DL, fuzzy logic, and optimization algorithms. Implementing the proposed method is very complicated because the CNN-AE model, IT2FR, and GWO methods have different adjustment parameters. Also, the proposed method has a long execution time, which is another limitation.

## 5. Discussion, Conclusion, and Future works

Brain disorders such as SZ and ADHD are among the most important brain diseases that seriously threaten many people's health worldwide [7][43]. ADHD and SZ disorders have many differences and similarities. In some studies, the presence of ADHD symptoms has been reported in people who are diagnosed with SZ in adulthood [115-116]. Also, ADHD is diagnosed in many children at genetic risk for SZ [128-130]. People with SZ in early adolescence often have symptoms of other mental disorders such as ADHD [128-130]. A diagnosis of ADHD in childhood may be a better predictor of adult SZ. For example, the risk of SZ in children with ADHD is approximately 4.3 times higher than in other adults [128-130]. The co-occurrence of ADHD and SZ may be due to shared genetic factors. ADHD and SZ may also have similarities in clinical treatment [106-107]. ADHD is often diagnosed during childhood, while SZ is often diagnosed in the 20s or 30s [115-116]. ADHD is treated with behavioral therapy methods, while SZ can be controlled with cognitive and emotional therapy (CET) [131]. ADHD does not require medication to treat it, but doctors may prescribe a stimulant to help the patient to focus [132-133]. SZ can be treated with antipsychotic drugs [115-116].

ADHD and SZ negatively affect brain function. So, various diagnostic methods have been introduced by specialist physicians to treat them [25]. Currently, the most used methods for the diagnosis of SZ and ADHD disorders include psychological tests and neuroimaging techniques [7] [43]. fMRI modalities are one of the best functional neuroimaging modalities, and physicians and doctors can use them to diagnose brain disorders [25]. Diagnosis of SZ and ADHD disorders using fMRI modalities is challenging. To overcome this, there has been significant growth over the years in research to diagnose various brain disorders, including SZ [6-7], ADHD [43,], etc. using AI (ML and DL) techniques.

In this paper, we presented a new computer-aided diagnosis system (CADS) based on the proposed CNN-AE architecture and IT2FR-GWO classifier for diagnosing SZ and ADHD disorders using rs-fMRI data. The steps of the proposed method comprise the dataset, preprocessing, feature extraction, and classification. First, the UCLA dataset [44] is used for experiments. As mentioned earlier, this dataset has rs-fMRI modality from HC, SZ, ADHD, and BD patients. We used the HC, SZ, and ADHD data for our experiments. In the second step, the rs-fMRI modality is preprocessed using FSL toolbox [45]. Then, we extracted the correlation coefficients matrices with $118 \times 118$ size as 2D images from rs-fMRI data of each subject for post-processing.

The third step is dedicated to feature extraction, which is performed by a proposed 2D CNN-AE architecture. In this step, the 2D images are fed to the input of the proposed CNN-AE model. The first novelty of this paper is an improved CNN-AE architecture. Figure (9) and Table (1) show that the proposed CNN-AE architecture has 7 CNN layers.

The last step of the proposed method is classification. A few ML-based, fuzzy type-1, and fuzzy type-2 methods are used for the classification of CNN-AE features. The DT [57], MLP [46], KNN [47], SVM [48], and RF [49] are ML classifier methods. Also, Type-1 fuzzy classification techniques include ANFIS [50], ANFIS-GA, ANFIS-POS, and ANFIS-GWO. The IT2FR, IT2FR-GA, IT2FR-PSO, and IT2FR-GWO are type-2 fuzzy classifier techniques. In this work, the IT2FR method was first introduced by the authors and is considered the most important novelty of the paper. In addition, another novelty of this paper is combining the IT2FR and GWO methods for classification. It should be noted that in the classification section, the K-Fold cross-validation with k = 10 is also used. The experiment results of the proposed method show that the use of the CNN-E model and the IT2FR-GWO method has led to increased accuracy in diagnosing SZ and ADHD disorders. Our findings suggest that significant different functional connectivity exists within these regions which are in line with previous literature [135-138].

In Table (8), the proposed method results are compared with related works in the diagnosis of SZ and ADHD using fMRI modalities. Table (8) shows that the researchers used different MRI datasets for SZ and ADHD detection. In related works, researchers have identified SZ from HC. However, our paper uses the UCLA dataset; the classes presented to the classifier in this paper include HC, SZ, and ADHD. Table (8) shows that this paper's proposed method has higher accuracy than other related works.

The proposed method of this paper can help physicians accurately diagnose SZ and ADHD from rs-MRI data. Also, the proposed method can be used in the future as application software for diagnosing SZ or ADHD in hospitals or specialized medical centers. Furthermore, in future works, new DL models such as transformers [100-101], attention mechanism [102-103], graph theory [104-105], etc., can be used for the diagnosis of SZ and ADHD.

Table 8. Comparison of the proposed method with past works.

| Works | Dataset | Modality | Type of Classes | Feature Extraction | Feature Selection | Classifier | Acc (%) |
|---|---|---|---|---|---|---|---|
| [75] | Clinical | T-fMRI | HC and SZ | Multivariate Connectome Features | Chi-Squared Test | SVM | Acc=71.6 |
| [76] | COBRE | rs-fMRI | HC and SZ | Different Graph Theoretical Features | -- | SVM | Acc=65 |
| [77] | COBRE | rs-fMRI, sMRI | HC and SZ | Dynamic Functional Connectivity | ENR | LR | Acc=71 |
| [78] | Multi-Site | rs-fMRI, T-fMRI | HC and SZ | Different Features | -- | Backus-Gilbert | Acc=58.6 |
| [79] | COBRE NMorphCH NUSDAST | sMRI | HC and SZ | Local Grey Matter Volume | LASSO | Enet-TV | Acc=68 |
| [80] | Clinical | rs-fMRI | HC and SZ | Brain-wide Seed Based Voxel-Wise Analysis | SelectFdr | GBDT | Acc=72.28 |
| [81] | Clinical | rs-fMRI | HC and SZ |  | T-Test | Majority Voting | Acc=73 |
|  |  |  |  |  | PCA |  |  |

| | | | | Spatial–Temporal Reconstruction Based on the ICA | Fisher | | |
|---|---|---|---|---|---|---|---|
| [82] | Clinical | sMRI | HC and SZ | Structure's Brain Volumes | -- | MLDA | Acc=73 |
| [83] | Different | rs-fMRI | HC and SZ | Demographic and Clinical Features | PCA | FCM | Acc=73 |
| [84] | COBRE | rs-fMRI, sMRI | HC and SZ | FCM | Mann–Whitney U test | SVM | Acc=69 |
| [85] | COBRE | rs-fMRI | HC and SZ | Consensus Functional Connections with High Discriminative Power | T-Test | LDA | Acc=76.34 |
| [86] | NUSDAST IMH | sMRI | HC and SZ | Inception ResNet | -- | SVM | Acc=70.98 |
| [87] | SchizConnect | sMRI | HC and SZ | 3D-CNN | -- | Softmax | Acc=70 |
| [88] | OpenfMRI | rs-fMRI | HC and SZ | GAN | -- | Different Methods | Acc=71.3 |
| [89] | COBRE UCLA | rs-fMRI | HC and SZ | Weighted Deep Forest | -- | Softmax | Acc=61 |
| [90] | COBRE | rs-fMRI | HC and SZ | DNN | -- | Softmax | Acc=77.8 |
| [91] | ADHD-200 | rs-fMRI | HC and ADHD | 1D-CNN | -- | Softmax | Acc=71.3 |
| [92] | ADHD-200 | rs-fMRI | HC and ADHD | DBN | -- | Softmax | Acc=69.83 |
| [93] | ADHD-200 R-fMRI Maps Project | rs-fMRI | HC and ADHD | 3D-CNN | -- | -- | Acc=65.67 |
| [94] | Different Datasets | rs-fMRI | HC and ADHD | DBN | -- | SVM | Acc=72.72 |
| [95] | ADHD-200 | rs-fMRI | HC and ADHD | DBN | -- | Softmax | Acc=72.73 |
| [96] | ADHD-200 | rs-fMRI | HC and ADHD | 3D-CNN | -- | Softmax | Acc=69.15 |
| [97] | ADHD-200 | rs-fMRI | HC and ADHD | DCNN | -- | Sigmoid | Acc=71.30 |
| [98] | ADHD-200 | rs-fMRI | HC and ADHD | SC-CNN-Attention | -- | Softmax | Acc=68.60 |
| [99] | ADHD-200 | rs-fMRI | HC and ADHD | C4D Network | -- | Softmax | Acc=70.4 |
| **Proposed Method** | **UCLA** | **rs-fMRI** | **HC, SZ, and ADHD** | **CNN-AE** | **--** | **IT2FR-GWO** | **Acc=72.71** |